\documentclass[conference]{IEEEtran}
\IEEEoverridecommandlockouts
\usepackage{cite}
\usepackage{amsmath,amssymb,amsfonts}
\usepackage{graphicx}
\usepackage{textcomp}
\usepackage{color,soul}
\usepackage[ruled,vlined,linesnumbered]{algorithm2e}
\usepackage[table,xcdraw]{xcolor}
\usepackage{multirow}
\usepackage{enumitem}
\usepackage{booktabs}
\usepackage{bm}
\usepackage{url}
\usepackage{makecell}

\def\BibTeX{{\rm B\kern-.05em{\sc i\kern-.025em b}\kern-.08em
    T\kern-.1667em\lower.7ex\hbox{E}\kern-.125emX}}

\begin{document}

\title{Simultaneously Evolving Deep Reinforcement Learning Models using Multifactorial Optimization
}

\author{
\IEEEauthorblockN{Aritz D. Martinez\IEEEauthorrefmark{2}\IEEEauthorrefmark{1},
Eneko Osaba\IEEEauthorrefmark{2}\IEEEauthorrefmark{1},
Javier Del Ser\IEEEauthorrefmark{2}$^,$\IEEEauthorrefmark{3} and
Francisco Herrera\IEEEauthorrefmark{4}}
\IEEEauthorblockA{\IEEEauthorrefmark{2}TECNALIA, Basque Research and Technology Alliance (BRTA), 48160 Derio, Bizkaia, Spain\\
Email: \{aritz.martinez, eneko.osaba, javier.delser\}@tecnalia.com}
\IEEEauthorblockA{\IEEEauthorrefmark{3}University of the Basque Country, 48013 Bilbao, Bizkaia, Spain\\}
\IEEEauthorblockA{\IEEEauthorrefmark{4}DaSCI Andalusian Institute of Data Science and Computational Intelligence. University of Granada. 18071 Granada, Spain\\}
\IEEEauthorblockA{\IEEEauthorrefmark{1} Corresponding authors. These authors have equally contributed to the work presented in this paper.}}
\maketitle

\begin{abstract} \label{sec:abstract}
In recent years, Multifactorial Optimization (MFO) has gained a notable momentum in the research community. MFO is known for its inherent capability to efficiently address multiple optimization tasks at the same time, while transferring information among such tasks to improve their convergence speed. On the other hand, the quantum leap made by Deep Q Learning (DQL) in the Machine Learning field has allowed facing Reinforcement Learning (RL) problems of unprecedented complexity. Unfortunately, complex DQL models usually find it difficult to converge to optimal policies due to the lack of exploration or sparse rewards. In order to overcome these drawbacks, pre-trained models are widely harnessed via Transfer Learning, extrapolating knowledge acquired in a source task to the target task. Besides, meta-heuristic optimization has been shown to reduce the lack of exploration of DQL models. This work proposes a MFO framework capable of simultaneously evolving several DQL models towards solving interrelated RL tasks. Specifically, our proposed framework blends together the benefits of meta-heuristic optimization, Transfer Learning and DQL to automate the process of knowledge transfer and policy learning of distributed RL agents. A thorough experimentation is presented and discussed so as to assess the performance of the framework, its comparison to the traditional methodology for Transfer Learning in terms of convergence, speed and policy quality , and the intertask relationships found and exploited over the search process.
\end{abstract}

\section{Introduction} \label{sec:intro}

Since its inception in the literature mainstream, Deep Learning has been typically used for modeling (i.e. generative distribution learning and predictive approaches) and actuation (correspondingly, Reinforcement Learning, RL). While much attention has been traditionally paid to the former, the irruption of multi-layered complex neural architectures to the RL paradigm marked a milestone in regards to the capability of artificial agents learn to optimally interact with a given environment. Among them, Deep Q Learning (DQL) introduced the use of Deep Neural Networks to predict the total reward expected to be gained after a particular action was taken on the environment. By leveraging this increased reward modeling capability along with other improvements (e.g. experience replay), DQL has since then excelled at learning over challenging RL domains, in some of them attaining near-human levels of performance.

Since the evaluation budget to solve a RL task is often assumed to be unlimited, the complexity of the environment does not have much relevance, given that the algorithm guarantees convergence to a sub-optimal solution. In this context, it has been proven that Deep Reinforcement Learning can achieve impressive results with a proper number of evaluations. In recent works such as the \texttt{HideAndSeek} game from OpenAI Labs \cite{baker2019emergent}, billions of evaluations are needed to make interesting behaviors emerge among the agents. Nevertheless, in such complex environments, or even in contexts where the amount of evaluations is limited, Transfer Learning is instead used to exploit the knowledge captured by models trained beforehand for other related tasks. This transferred knowledge is adopted as the initial state for the agent to be learned in the target scenario, which allows reducing the time needed to grasp and acquire sophisticated behavioral skills. 

Although Transfer Learning has a straightforward application among Deep Learning models, its effectiveness is stringently subject to the similarity between origin and destination tasks \cite{yosinski2014transferable}. Unfortunately, it is not always the case in which two tasks can be claimed to be similar to each other and suitable for Transfer Learning before actually testing their empirical performance. For instance, in the \texttt{HideAndSeek} scenario tackled in \cite{baker2019emergent}, Transfer Learning is used to make agents perform different complex tasks departing, as starting state of the Deep RL model, knowledge captured for other tasks such as \texttt{ObjectCounting}, \texttt{LockAndReturn}, \texttt{ShelterConstruction} or \texttt{SequentialLock}, among others (i.e. pretrained on these tasks). Despite the benefits of Transfer Learning for the RL realm, determining the transferability of knowledge among tasks is, to date, delegated to common sense and intuition held on the scenarios at hand.

The work presented in this paper aims to take a step forward in this direction. Specifically, we propose a framework for simultaneously learning multiple DQL models for RL tasks, leveraging the existing relationships and complementarities among such tasks during the training phase. To this end, the learning process is formulated as an optimization problem defined on a common search space, from which solutions to specific DQL models for every task can be decoded. Efficiently solving this optimization problem is done by means of Transfer Optimization, an emerging paradigm for tackling different problems by automatically exploiting synergies among them learned over the optimization process \cite{gupta2017insights}. Several categories can be established under this paradigm (e.g. \emph{sequential transfer} \cite{feng2015memes}). In this work we select \textit{multitasking} to dynamically optimize multiple DQL models, and to capitalize on the relationships between RL scenarios during the search \cite{wen2017parting,gupta2016genetic}. In particular, we resort to the so-called Multifactorial Evolutionary Algorithm (MFEA, \cite{gupta2015multifactorial}), a specific multitasking algorithm that relies on evolutionary search operators (also known as Evolutionary Multitasking \cite{Ong2016}). The intrinsic knowledge transfer capability between candidates featured by MFEA allows for a simultaneous co-evolution of multiple DQL models, achieving faster convergence and better performance levels by virtue of similarities among RL tasks. 

A experimental setup is devised to analyze and validate the performance of the proposed framework when undertaking several simple RL scenarios. As discussed from the obtained results, MFEA not only renders an excellent performance when learning all DQL models at the same time in terms of accuracy and convergence speed, but also evinces whether the search algorithm identifies and exploits the inherent relationships between the defined RL scenarios. The results stemming from this research work can buttress the idea that a unique search process can face the main drawbacks of DQL: i) on the one hand, sharing a unique search space avoids the evolution process to get stuck in local optima; and ii) on the other hand, knowledge is shared between different tasks without any information on whether they are related anyhow, hence acting as an \emph{adaptive} transfer learning mechanism.

The rest of the paper is organized as follows: first, Section \ref{sec:background} introduces key concepts on MFEA, DQL and Transfer Learning that are required for a proper understanding of the present contribution. Next, Section \ref{sec:approach} describes in detail the proposed framework. Experiments and results are presented and discussed in Section \ref{sec:experimentation}. Finally, conclusions and prospects towards future research lines are given in Section \ref{sec:conclusions}.

\section{Background} \label{sec:background}

As stated above, this section introduces background concepts related to the present work. To begin with, Subsection \ref{sec:mfea} formally exposes the MFEA algorithm adopted in this work. Then, Subsection \ref{sec:dl} provides an introduction to DQL, while Subsection \ref{sec:tl} explains how Transfer Learning is usually performed on RL scenarios.       

\subsection{Multifactorial Evolutionary Algorithm} \label{sec:mfea}

MFEA is a recently proposed evolutionary algorithm for addressing multiple optimization problems using bio-cultural schemes of multifactorial inheritance \cite{gupta2015multifactorial}. Briefly explained, MFEA considers an scenario with $K$ optimization problems (\emph{tasks}), defined as $\{T_1, \ldots, T_K\}=\{T_k\}_{k=1}^K$. Without loss of generality we assume that these tasks are minimization problems, each defined over a search space $\bm{\mathcal{X}}_{k}$ and $D_k=|\bm{\mathcal{X}}_{k}|$ (dimensionality). The goal is to find solutions $\{\mathbf{x}_k^\ast\}_{k=1}^K$ such that $\mathbf{x}_k^\ast=\arg_{\mathbf{x}\in\bm{\mathcal{X}}_{k}}\min F_k(\mathbf{x})$ $\forall k\in\{1,\ldots,K\}$, where $F_{k}:\bm{\mathcal{X}}_{k} \mapsto \mathbb{R}$ is the objective function of task $T_k$.

Some concepts must be posed before continuing further. Firstly, it is necessary to define a unified search space $\bm{\mathcal{X}}^U$ for all tasks, wherein solutions to each of such problems can be encoded and decoded. The form taken by this unified search space $\bm{\mathcal{X}}^U$ depends roughly on the optimization problems to be solved, with several alternatives contributed to date (e.g. Random Keys \cite{Bean:94}). The dimensionality of the unified search space is usually\footnote{This depends on how solutions belonging to the unified search space are decoded to solutions of each problem $T_k$.} set to be the maximum among the dimensions of the tasks under consideration, i.e. $|\bm{\mathcal{X}}^U| = \max_k D_k$. 

In order to be able to compare among candidate solutions in the unified space, it is necessary to establish a global search method. According to evolutionary computation principles, a population of candidates $\{\mathbf{x}^p\}_{p=1}^P$ is defined, wherein each individual $\mathbf{x}^p$ can be decoded to $\mathbf{x}_k^p\in\bm{\mathcal{X}}_k$ related to task $T_k$. Once decoded, several concepts are defined to rank and evolve the individuals within the population:
\begin{itemize}[leftmargin=*]
    \item Factorial cost ($\Psi_{k}^p\in\mathbb{R}$): The factorial cost of candidate $\mathbf{x}^p$ in task $T_k$ is given by $\Psi_{k}^p=\lambda \cdot \delta_{k}^p + F_{k}(\mathbf{x}_k^p)$, where $\lambda$ is a penalizing multiplier, $\delta_{k}^p$ represents the total constraint violation of $\mathbf{x}^p$ at task $T_k$, and $F_{k}(\mathbf{x}_k^p)$ denotes the objective value once $\mathbf{x}^p$ has been decoded to $\mathbf{x}_k^p$. If there is not constraint violation, $\Psi_{k}^p = F_{k}(\mathbf{x}_k^p)$.       

    \item Factorial rank ($r_{k}^p\in\{1,\ldots,P\}$) is defined as the position or rank of candidate $\mathbf{x}^p$ in task $T_k$ after sorting the population in ascending order of $\Psi_{k}^p$.
    
    \item Scalar fitness ($\varphi^p\in\mathbb{R}[0,1]$) is given by the inverse of the factorial rank $r_{k}^p$ of candidate $\mathbf{x}^p$ in the task in which it performs best, i.e., $\varphi^p = 1/ \left(\min_{k \in \{1...K\}}r_{k}^p\right)$.
        
    \item Skill factor ($\tau^p\in\{1,\ldots,K\}$) is the task in which $\mathbf{x}^u$ performs best, i.e., $\tau^p = \arg_{k\in\{1,\ldots,K\}} \min r_k^p$.     
\end{itemize}

When the scalar fitness $\varphi^p$ has been computed $\forall \mathbf{x}^p$, they can be compared to each other by assuming that $\mathbf{x}^p$ dominates $\mathbf{x}^{p\prime}$ if $\varphi^p > \varphi^{p\prime}$. Based on this relationship of dominance and the encoding/decoding procedure from $\bm{\mathcal{X}}^U$ to each $\bm{\mathcal{X}}_k$ and vice versa, an evolutionary optimization algorithm can be defined, comprising crossover, mutation and selection operators operating on the aforementioned population of individuals.
\begin{algorithm}[tbh!]
	\label{alg:implementedMFEA}
	\caption{Multifactorial Evolutionary Algorithm}
	\SetAlgoLined
	\DontPrintSemicolon
	Initialize at random a population $\{\mathbf{x}^p\}_{p=1}^P$\;
	Compute factorial costs $\Psi_k^p$ $\forall k,p$\;
	Assign a skill factor $\tau^p$ to each $\mathbf{x}^p$\;
	\While{termination criterion not reached}{
		Apply operators to $\{\mathbf{x}^p\}_{p=1}^P$, yielding $\{\mathbf{x}^{p,\vartriangle}\}_{p=1}^{P^\vartriangle}$\;
		Evaluate each $\mathbf{x}^{p,\vartriangle}$ on its parents' skill factor\;
		Merge $\{\mathbf{x}^{p,\vartriangle}\}_{p=1}^{P^\vartriangle}$ and $\{\mathbf{x}^{p}\}_{p=1}^P$\;
		Update $\varphi^p$ and $\tau^p$ of merged candidates\;
		Select $P$ best candidates in terms of \;
	}
	Return the best candidate for each of the $K$ tasks\;	
\end{algorithm}

This is indeed the approach followed by MFEA, whose search procedure is described in Algorithm \ref{alg:implementedMFEA}. Apart from the concept of unified search space and the selection based on factorial ranks and factorial costs, MFEA differs from standard evolutionary heuristics also in terms of search operators, featuring novel methods (namely, \textit{Assortative Mating} and \textit{Vertical Cultural Transmission}) that favor the exchange of knowledge among tasks. On one hand, \textit{Assortative Mating} (Algorithm \ref{alg:assortative_mating}) imposes that candidates with the same skill factor mate more likely than two candidates excelling at two different tasks. 
\begin{algorithm}[tbh!]
	\label{alg:assortative_mating}
	\caption{Assortative mating in MFEA} 
	\DontPrintSemicolon
	\SetAlgoLined
	Define a threshold $\varphi \in \mathbb{R}[0,1]$\;
	Draw a random number $r \sim \text{Uniform}(0,1)$\; 
	Select two parents $\mathbf{x}^{p_1}$ and $\mathbf{x}^{p_2}$ from the population\;
	\uIf{$\tau^{p_1}=\tau^{p_2}$ \normalfont{\textbf{or}} $r<\varphi$}{
		Apply crossover and mutation on $\mathbf{x}^{p_1}$ and $\mathbf{x}^{p_2}$ to generate two offspring $\mathbf{x}^{p_1,\vartriangle}$ and $\mathbf{x}^{p_2,\vartriangle}$\;
	}
	\Else{
		Apply mutation on $\mathbf{x}^{p_1}$ to yield $\mathbf{x}^{p_1,\vartriangle}$\;
		Apply mutation on $\mathbf{x}^{p_2}$ to yield $\mathbf{x}^{p_2,\vartriangle}$\;
	}
	
\end{algorithm}

On the other hand, \emph{Vertical Cultural Transmission} enforces that candidates are not evaluated in all optimization tasks $\{T_1,\ldots,T_K\}$, but only on one of their parents' skill factors. This specific aspect makes MFEA competitive in terms of computational complexity. In addition to these two search operators, MFEA implements a local search method before evaluating a candidate, which we will not consider in our study for reasons disclosed later. Finally, once the produced offspring have been evaluated and merged with those from the previous generation, selection based on scalar fitness $\varphi^p$ is applied, by which the best $P$ individuals in terms of this fitness are retained in the population for the next generation.

\subsection{Deep Reinforcement Learning} \label{sec:dl}

As mentioned in the introduction, the proposed framework deals with distributed RL problems undertaken by means of DQL models, whose learning process is formulated as an optimization problem and solved jointly via evolutionary multitasking. DQL has gained a growing popularity during recent years for their superior modeling \cite{mnih2013playing} and multi-agent cooperation capabilities \cite{baker2019emergent}. However, what makes DQL models particularly interesting for the purpose of this study is their suitability to be trained by using metaheuristic algorithms\cite{such2017deep}, showing competitive results when compared to traditional back-propagation. For the sake of completeness, we next elaborate briefly on the concepts underneath DQL.

In general, the goal of RL is to learn a policy, usually denoted as $\pi_{\bm{\theta}}(s)$ (with ${\bm{\theta}}$ denoting the parameters defining the policy itself), to efficiently perform a task depending on a state $s\in\mathcal{S}$. In general terms, a RL approach is described by an agent in charge of taking an action $a_t=\pi_{\bm{\theta}}(s_t)$ from a set of actions $\mathcal{A}$, every time a new state $s_t\in\mathcal{S}$ is presented to it. The environment, where the task is defined, returns a reward of performing action $a_t$ in state $s_t$, and informs the agent with the next state $s_{t+1}$ of the environment. The search for a policy $\pi_{\bm{\theta}}(s)$ capable of maximizing this reward and optimally solving the RL task is not trivial, unchaining a flurry of different algorithmic approaches and heuristics for this purpose \cite{8103164}. 
\begin{figure}[tbh!]
	\centering
	\includegraphics[width=0.8\linewidth]{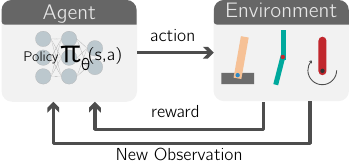}
	\caption{General architecture of DQL, in which the policy to be learned is implicitly embedded in the trainable parameters of a Deep Neural Network.}
	\label{fig:dql}
\end{figure}

One of such different policy learning methods is DQL, in which the policy is represented by the weights and biases of a Deep Neural Network (Figure \ref{fig:dql}). The $Q$ in DQL comes from the traditional $Q$ learning, and refers to the long-term return $Q(s_t, a_t)$ predicted for every state-action pair at instant $t$. This value is updated every time the agent takes an action, as per:
\begin{equation}
Q(s_t, a_t)\hspace{-0.7mm}\leftarrow\hspace{-0.7mm}Q(s_t, a_t) + \alpha [ R_{t} + \gamma \max_{a\in \mathcal{A}} Q(s_{t+1}, a)\hspace{-0.5mm}-\hspace{-0.5mm}Q(s_t, a_t)],\nonumber
\end{equation}
where $Q(s_t,a_t)$ denotes the current $Q$ value associated to taking the action $a_t$ at state $s_t$, $R_t$ is the reward received when moving from state $s_t$ to state $s_{t+1}$, $\alpha\in\mathbb{R}[0,1]$ is a learning rate, and $\gamma$ is a discount factor that lowers the contribution of future rewards in the computation. As opposed to this update rule, DQL resorts to a Deep Neural Network to approximate the $Q$ value associated to every task (outputs) given the current state (input). As a result, the need for computing and updating $Q$ for every (state, action) pair is overridden, making it suitable for RL problems of high dimensionality.

Since the learned policy $\pi_{\bm{\theta}}(s)$ is implicitly encoded in the weights and biases of the Deep Neural Network (${\bm{\theta}}$), the behavior of the agent when acting on the environment only depends on the values of such neural parameters. To set optimally these values, gradient descent is often utilized\footnote{Due to the lack of space we skip specific aspects of DQL model training such as experience replay, target network and reward clipping.} to minimize a $Q$-dependent squared error loss function with respect to ${\bm{\theta}}$:
\begin{equation}\label{eq:dql_loss}
\min_{\bm{\theta}} \frac{1}{2} \left(M_{\bm{\theta}}(s_t, a_t) - [R_{t} + \gamma \max_{a\in \mathcal{A}} M_{\bm{\theta}}(s_{t+1}, a)]\right)^2,
\end{equation}
where $M_{\bm{\theta}}(s,a)$  the output of the neural network with parameters ${\bm{\theta}}$ corresponding to action $a$ when observing state $s$. 

Indeed, the training problem can be seen as an optimization task wherein parameter values act as decision variables, and the loss function as the objective to be minimized. This fact underlies the reason why metaheuristic algorithms have been proposed to evolve DQL models, with the goal of harnessing the highly parallel nature of these solvers towards finding good policies at a lower computational cost \cite{such2017deep,igel2003neuroevolution,khadka2018evolutionary,salimans2017evolution}. More recently, alternative techniques such as Novelty Search have been hybridized with metaheuristic methods to reach even better results for evolving DQL models \cite{conti2018improving}. Neuroevolution for DQL models has been approached with evolutionary algorithms as well \cite{koutnik2014evolving}.

\subsection{Transfer Learning} \label{sec:tl}

Finally, a third ingredient involved in our framework is Transfer Learning, which refers to the knowledge transfer between models solving tasks that share some similarities \cite{glatt2016towards}. Conforming to intuition, the most closely tasks are related to each other, the most effective the transfer of knowledge between models will be. This statement also holds in Transfer Optimization, in which the relationship between the problems being solved also drives the potential benefit of exchanging knowledge among them. The main rationale of Transfer Learning when utilized in the context of data modeling tasks is to quickly adapt a model for performing well in different environments, with a lower demand for training data or at a lower training cost. 

While Transfer Learning has been a thoroughly addressed matter of study, the simplicity by which Transfer Learning can be realized between Deep Learning models has allowed them to dominate this research arena, particularly in image recognition and reinforcement learning scenarios \cite{taylor2009transfer,shao2014transfer}. The most common approach is to reuse pre-trained models learned for the source task(s) as a initialization point of the model of the target task. To this end, the most straightforward approach is to copy a fraction of the parameter values of the pre-trained model to the model that faces the target task. Once this is done, some layers of the model of the target task are frozen, whereas parameter belonging to the rest of the layers are trained via back-propagation. 

Despite this apparent simplicity, it is not trivial to determine how many layers/parameters to transfer between models: the more similar the tasks are, the higher the number of parameters that can be transferred to the model of the target task will be. Furthermore, it still remains unclear whether the degree of similarity between origin and target task can be easily quantified in complex modeling problems, specially in those where no intuition can be held in this matter. In addition, the mapping between the degree of similarity among tasks and the amount of parameter values transferred between models is still open to further research. With all this been said, Transfer Learning can be useful between DQL to develop general controllers that can adapt themselves to different environments, and solve accurately RL problems therein. The framework presented in the next section aims precisely at this goal, further complementing it with its capability to enforce information transfer among RL tasks during the learning process.

\section{Proposed Framework} \label{sec:approach}

The framework proposed in this paper simultaneously evolves DQL models aimed at solving distinct albeit interrelated RL tasks. To perform this efficiently, we adopt MFEA as the method in charge of solving the different task bunch. Besides, a novel encoding scheme is devised, whose design is inspired by the way Transfer Learning is implemented among neural networks.

To begin with, the optimization problems $\{T_1,\ldots,T_K\}$ represent RL tasks defined on a set of different environments $\mathcal{E}=\{E_1,\ldots,E_N\}$, such that the environment index on which problem $T_k$ is defined is given by $\xi(k)$, with $\xi:\{1,\ldots,K\}\mapsto \{1,\ldots,N\}$. Furthermore, each of such environments can be \emph{configured} with a set of parameters $\beta$ that permit to tailor different configurations of the environment for the task to be solved. For instance, in a $\texttt{cartpole}$ game, $\beta$ could represent the gravity force or the length of the pole to be kept upright\footnote{Further information about these exemplifying games, which are also used in the experiments, can be found in the OpenAI Gym Library \cite{brockman2016openai}.}. Therefore, tasks to be solved may belong to different environments, wherein tasks defined on the same environment may operate under different configurations. Ideally, knowledge transfer should occur more likely among tasks defined on the same environment (\emph{intra-environment transfer}), as long as the configurations $\beta$ do not make the tasks radically different to each other. When extrapolating this statement to the knowledge exchange among different environments (corr. \emph{inter-environment transfer}), greater differences among the tasks can be expected, yet still leaving room for knowledge exchange (e.g. levering a pole in \texttt{cartpole} should contribute to swinging up a two-link robot in the \texttt{acrobot} game).

Automating this knowledge transfer among DQL models aimed to solve different tasks is what motivates the adoption of MFEA. To this end, for each task $k\in\{1,\ldots,K\}$ a DQL model $M_{\bm{\theta}_k}^k$ is sought, whose parameters $\theta_k$ (weights, biases) constitute the local decision variables that MFEA aims to optimize via a single exploration over a unified search space. Therefore, according to the notation introduced for MFEA in Subsection \ref{sec:mfea}, $\bm{\theta}_k\equiv \mathbf{x}_k$ and $\bm{\mathcal{X}}_k$ is given by $\mathbb{R}^{D_k}$, with $D_k=|\bm{\theta}_k|$. Based on this notation, the problem $T_k$ to be solved is formulated as the maximization of the final reward of the DQL model $M_{\bm{\theta}_k}^k$ averaged over a number $N_{tst}$ of test episodes run over environment $\xi(k)$:
\begin{equation}
\text{Problem }T_k: \max_{\bm{\theta}_k\in \mathbb{R}^{D_k}} \frac{1}{N_{tst}}\sum_{t=1}^{N_{tst}} R(M_{\bm{\theta}_k}^k;E_{\xi(k)},t),
\end{equation}
where $R(M;E,t)$ denotes the final reward achieved by model $M$ on environment $E$ at test episode $t$. We note here that the above objective differs from the one pursued in the traditional DQL training approach, given in Expression \eqref{eq:dql_loss}. The reason is that we do not seek to learn $\bm{\varphi}_k$ online over epochs by interacting with the environment, but rather to optimize them off-line. Nevertheless, this clarification stimulates research lines further exposed in Section \ref{sec:conclusions}.

In order to realize a unified search space $\bm{\mathcal{X}^U}$ on which to perform the multifactorial search, two important factors must be taken into account when designing the encoding and decoding procedures between $\bm{\mathcal{X}^U}$ and $\bm{\mathcal{X}}_k$: i) they must be computationally efficient; and ii) they must harness any a priori knowledge on the type of tasks to be solved. Indeed, when Transfer Learning is applied to RL, there are some considerations that can be taken into account in order to obtain a better performance \cite{taylor2009transfer}. It is known that neural networks learn general patterns in their first layers, while the last ones retain specialized information about the task under consideration. Consequently, as mentioned in Subsection \ref{sec:tl}, Transfer Learning among Deep Neural Networks is mostly made by sharing network parameters corresponding to the first layers of the network, while delegating the specialization for the task to be solved to the parameters of the last layers of the neural architecture.

Bearing these observations in mind, it seems natural to embrace them when designing $\bm{\mathcal{X}^U}$. Specifically, we partition any given solution $\mathbf{x}^p\in\bm{\mathcal{X}^U}$ in two parts:
\begin{itemize}[leftmargin=*]
\item A first part $\bm{\mathcal{X}}_{sh}^U$ representing the parameters of the $L_{sh}$ first layers of all models $\{M_{\bm{\theta}_k}^k\}_{k=1}^K$, whose values will be the same across tasks; and
\item $K$ parts $\{\bm{\mathcal{X}}_k^U\}_{k=1}^K$, each embedding the parameters of the last layers of \smash{$M_{\bm{\theta}_k}^k$} that are specialized for $T_k$ and hence, not shared across different tasks.
\end{itemize}
\begin{figure}[h!]
	\centering
	\includegraphics[width=\columnwidth]{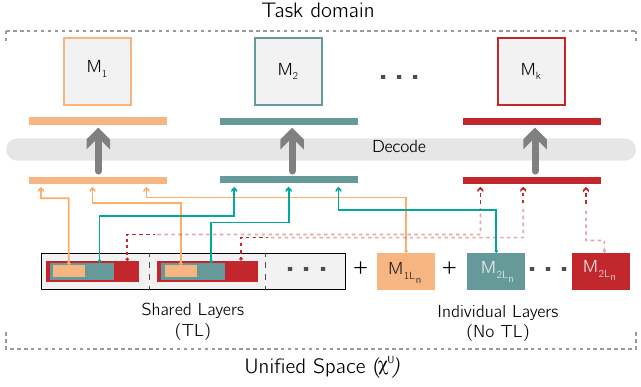}
	\caption{Encoding and decoding scheme of the proposed MFEA framework.}
	\label{fig:codification}
\end{figure}

The decoding process is graphically depicted in Figure \ref{fig:codification}, where it can be observed that candidate solutions $\bm{\theta}_k\equiv \mathbf{x}_k$ are built by extracting the shared weights and biases in $\bm{\mathcal{X}}_{sh}^U$, followed by those in $\bm{\mathcal{X}}_k^U$. Therefore, the dimensionality of the unified search space is given by:
\begin{align} \label{eq:U}
|\bm{\mathcal{X}}^U| &= \sum_{l=1}^{L_{sh}} \max_{k}\{|\bm{\varphi}_k|: \bm{\varphi}_k\in \text{layer $l$ of $M_{\bm{\theta}_k}^k$}\}\nonumber\\
&+ \sum_{k=1}^K\sum_{l=L_{sh}+1}^{L^k} |\bm{\varphi}_k|: \bm{\varphi}_k\in \text{layer $l$ of $M_{\bm{\theta}_k}^k$},
\end{align}
where $L^k$ denotes the number of layers of $M_{\bm{\theta}_k}^k$. Once $\bm{\mathcal{X}}^U$ has been defined, knowledge among $\{T_k\}_{k=1}^K$ is intrinsically shared via the unified search space and the MFEA evolutionary operators (Algorithm \ref{alg:implementedMFEA}). As such, the population of individuals $\{\mathbf{x}^p\}_{p=1}^P$ defined on the unified search space is evolved over generations by successively applying crossover and mutation operators through the assortative mating procedure of Algorithm \ref{alg:assortative_mating}. This operator favors the implicit genotype transfer between tasks, allowing a RL problems to benefit from the knowledge gained to solve other RL problems embedded in the weights and biases belonging to the shared partition $\bm{\mathcal{X}}_{sh}^U$ of the search space $\bm{\mathcal{X}}^U$. As per the evolutionary operators, we utilize, at no loss in generality, those prescribed in \cite{da2017evolutionary}: Simulated Binary Crossover (SBX) and polynomial mutation. 

Once all details of the proposed framework have been given, we are in a position to assess its performance via computer simulation. The next section elaborates and discusses on the results obtained therefrom.

\section{Experiments and Results} \label{sec:experimentation}

We proceed by evaluating the performance of the proposed MFEA-based framework when evolving multiple DQL models simultaneously. To this end, we structure the experiments and discussion to provide an informed answer to several research questions (RQ):
\begin{itemize}[leftmargin=*]
\item RQ1: Can MFEA optimize DQL models individually? Do they efficiently solve the considered RL tasks?
\item RQ2: Can MFEA optimize DQL models simultaneously, exploiting their synergies to achieve a better performance?
\item RQ3: Can we quantify the intra- and inter-environment knowledge transfer between RL tasks realized by MFEA? 
\end{itemize}

To elaborate on these questions, we first establish the $N=3$ RL environments $\{E_n\}_{n=1}^3$ involved in the experimentation. From the set of environments provided by OpenAI Gym Library we select two environments with a discrete action space ($E_1=\texttt{cartpole}$ and $E_2=\texttt{acrobot}$), and a third environment with continuous action space ($E_3=\texttt{pendulum}$). The configuration $\beta$ of these environments has been manipulated in order to yield a total of $K=12$ tasks $\{T_k\}_{k=1}^{12}$, which we will hereafter refer to as $\texttt{environment}(\alpha)$, with $\alpha\in\{A,B,C,D\}$ indicating the configuration set. Environments and configurations considered in the experiments are summarized in Table \ref{tab:new_envs}. 
\begin{table}[h!]
	\centering
	\caption{Environments and configurations of the experiments}
	\resizebox{\columnwidth}{!}{
		\begin{tabular}{lccccc}
			\toprule
			Environments & Configuration & $\alpha=A$ & $\alpha=B$ & $\alpha=C$ & $\alpha=D$ \\ 
			\midrule
			$\texttt{cartpole}(\alpha)$ & Pole length & 0.5 & 0.6 & 0.7 & 0.4 \\ 
			$\texttt{acrobot}(\alpha)$ & Joints' length & 1 & 1.2 & 1.4 & 1.6 \\ 
			$\texttt{pendulum}(\alpha)$ & Max speed/Max torque & 8 / 2.0 & 6 / 2.0 & 10 / 2.0 & 8 / 2.5 \\ \bottomrule
		\end{tabular}%
	}
	\label{tab:new_envs}
\end{table}

\begin{figure}[h!]
	\centering
	\includegraphics[width=0.65\linewidth]{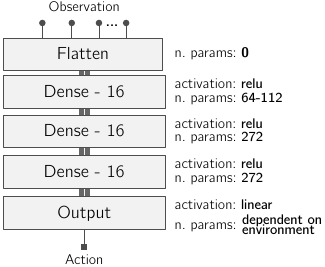}
	\caption{Deep Neural Network architecture and number of parameters of every layer in charge of solving the environments under consideration.}
	\label{fig:DQN_architecture}
\end{figure}

In order to solve the aforementioned tasks, the Deep Neural Networks $\{M_{\bm{\theta}_k}^k\}_{k=1}^K$ used for DQL are set with the same number of layers, parameters and overall architecture. Figure \ref{fig:DQN_architecture} (next page) depicts this common neural architecture, as well as the number of parameters on each layer. We further set $L_{sh}=3$ shared layers among the tasks, yielding a unified search space $\bm{\mathcal{X}}^U$ with dimensionality $|\bm{\mathcal{X}}^U|=962$ decision variables (weights and biases) as per Expression \eqref{eq:U}. As for the MFEA, it is relevant to emphasize that its configuration is kept fixed for all experiments: a population size of $P=100$ candidates, random initialization, assortative mating operator controlled by threshold $\varphi=0.3$ (see Algorithm \ref{alg:assortative_mating}), and stopping criteria given by a total of $60$ generations (namely, $6\cdot 10^3$ fitness evaluations). Furthermore, the fitness of candidates is computed as the average of the final reward achieved by the candidate over $N_{tst}=50$ episodes of the corresponding environment. 

To account for the statistical variability of the results, 5 independent runs of every experiment have been executed, reporting on performance statistics averaged over such runs. Each model, once optimized by the proposed framework, is tested on $250$ episodes of its corresponding task, from which the performance statistics mentioned previously are computed. All the tests conducted in this experimentation have been run on an Intel(R) Xeon W-2123 processor running at 3.6 GHz with 32 Gb RAM. The source code producing the results presented in what follows has been made available at \cite{martinez:19}.

\subsection*{RQ1: Can MFEA optimize DQL models individually? Do they efficiently solve the RL tasks?} \label{exp:rq1}

\begin{figure}[tbh!]
	\vspace{-1mm}
    \centering
    \includegraphics[width=0.8\columnwidth]{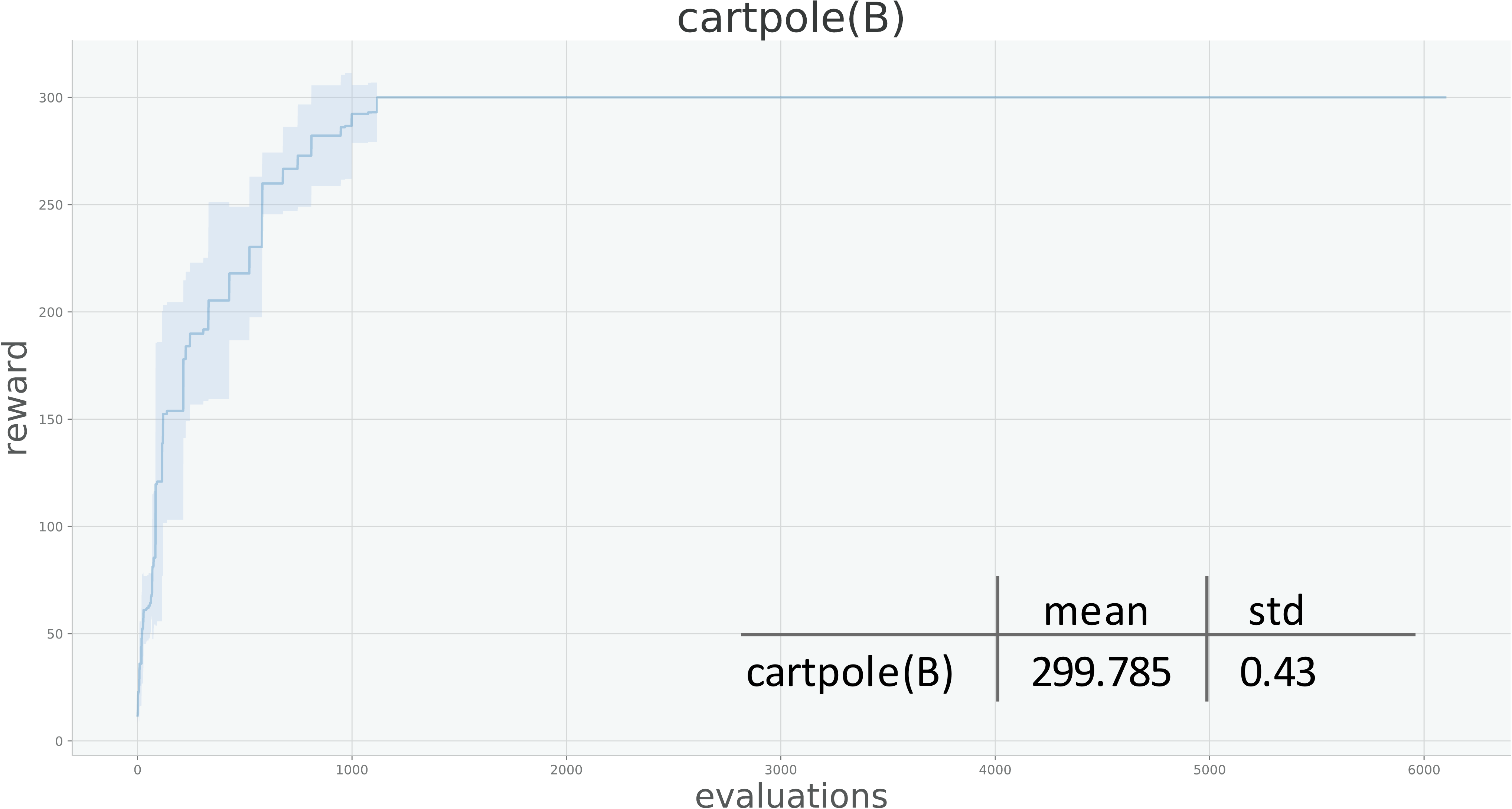} \\
    \vspace{3mm}
    \includegraphics[width=0.8\columnwidth]{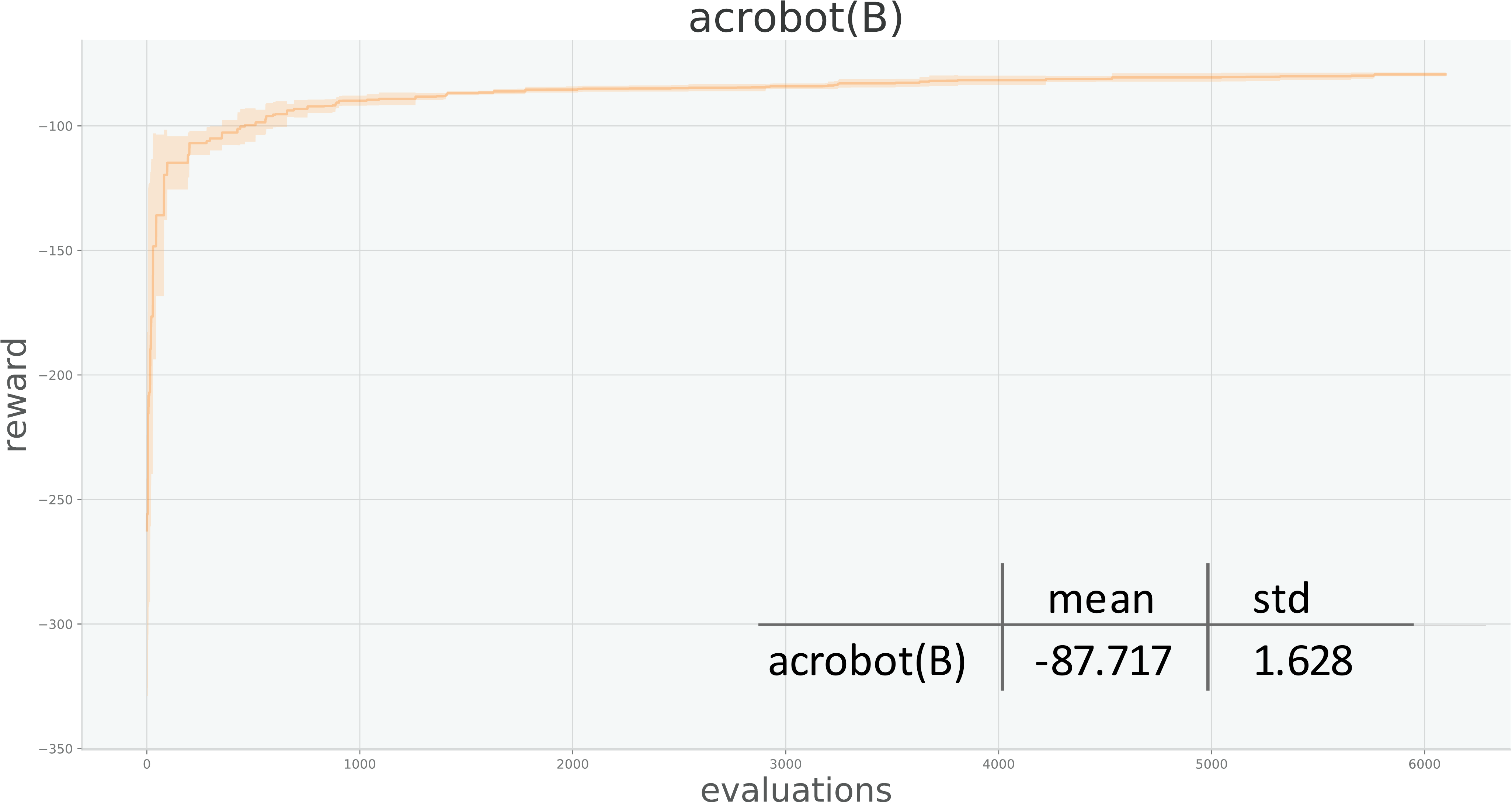} \\
    \vspace{3mm}
    \includegraphics[width=0.8\columnwidth]{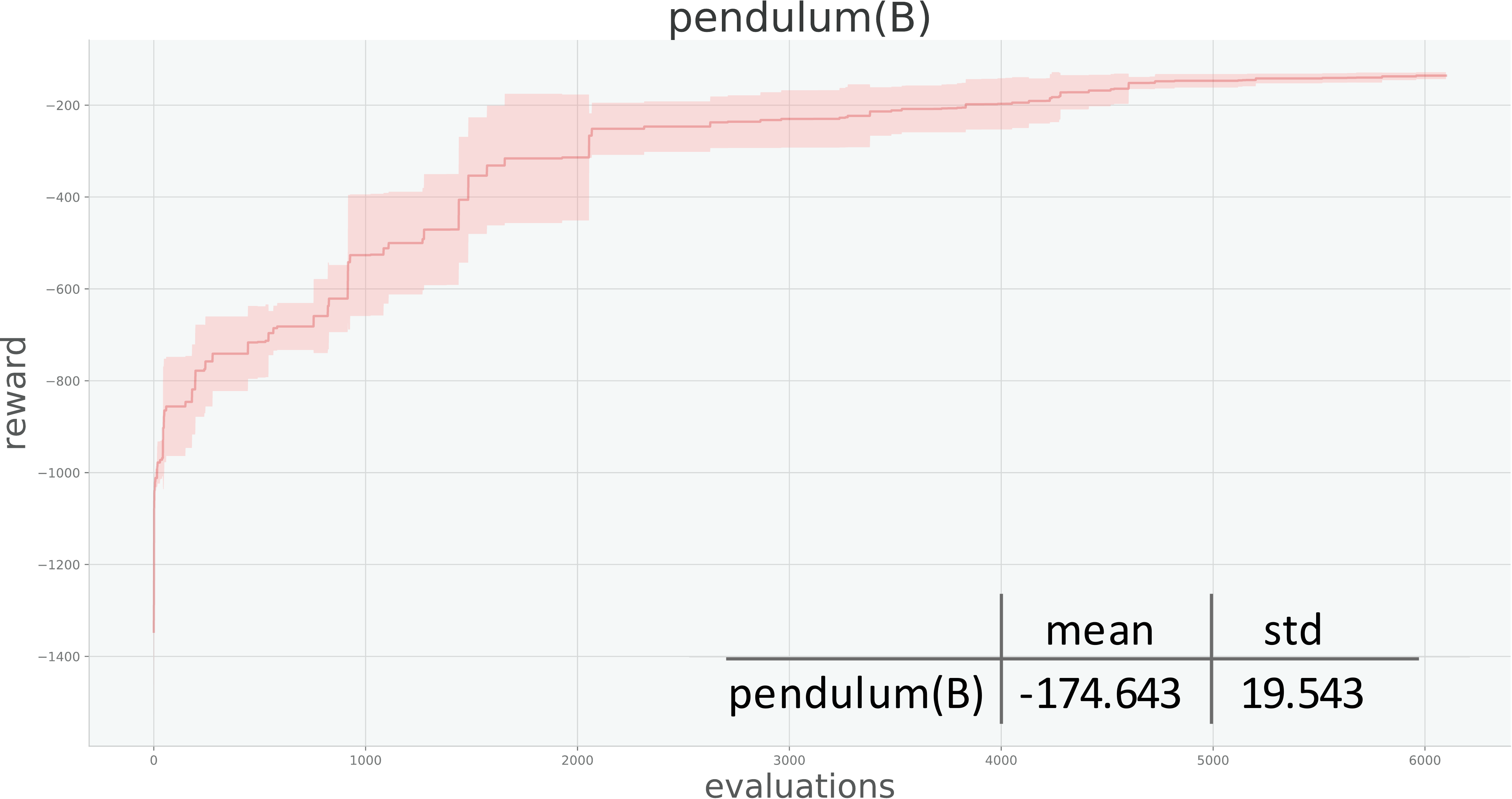} \\
    \caption{Evolution of MFEA when individually evolving DQL models aimed to solve tasks \texttt{cartpole}$(B)$, \texttt{acrobot}$(B)$ and \texttt{pendulum}$(B)$. Mean and standard deviation values of the evolved model tested on 500 episodes of their correspondent environment is shown.\vspace{-1mm}}
    \label{fig:OneEnvEv}
\end{figure}

First, MFEA is tested on each of the tasks individually, i.e. by just considering one single task ($K=1$). This is expected to yield the best models MFEA can achieve, setting the highest reward that can be reached on every task. Figure \ref{fig:OneEnvEv} exemplifies the evolution of the average reward attained by MFEA as a function of the number of evaluations for tasks \texttt{cartpole}$(B)$, \texttt{acrobot}$(B)$ and \texttt{pendulum}$(B)$. The shaded area represents the $\pm std$ range centered on the mean (bold line), computed over 5 independent runs and 250 test episodes per every run. 

Some interesting aspects can be drawn from these plots: to begin with, the evolutionary operators defined in MFEA succeed at converging fast and effectively towards well-performing models, characterized by the mean and standard deviation of their reward shown at the right bottom corner of each plot. In terms of convergence, \texttt{acrobot}$(B)$ and \texttt{cartpole}$(B)$ are able to reach the highest performance in approximately $10$ generations, whereas \texttt{pendulum}$(B)$ converges in around 20 generations. This relatively worse convergence featured by \texttt{pendulum}$(\alpha)$, which was indeed observed $\forall \alpha\in\{A,B,C,D\}$, will have a strong implication in the results obtained to gain insights on RQ3. 

\subsection*{RQ2:  Can MFEA optimize DQL models simultaneously, exploiting their synergies to achieve a better performance?} \label{exp:rq2}

We proceed by inspecting the multi-environment evolving capabilities of MFEA. A single population of solutions defined on the unified search space $\bm{\mathcal{X}}^U$ is shared between all tasks, while the evolution of each individual task is performed with the addition of the intrinsic knowledge transfer endowed by MFEA. The main goal of this experiment is to check whether MFEA is a good solver for multiple environments, giving quantitative metrics of the quality (reward) achieved by models evolved via a single search process. Therefore, we retrieve the reward statistics obtained for every task when the corresponding DQL model is evolved by MFEA in isolation with respect to the other tasks (RQ1). We then compare these scores to those obtained when all tasks corresponding to a given environment are jointly evolved, potentially promoting intra-environment knowledge transfer. Since the number of fitness evaluations is kept fixed to $60\cdot 10^3$ in all cases, the difference among such scores will evince the performance degradation when all models are jointly optimized via MFEA.
\begin{table}[h!]
	\centering
	\vspace{-4mm}
	\caption{Average and standard deviation of rewards obtained by MFEA on different environments when optimizing models separately or simultaneously (intra-environment transfer)}
	\label{tab:model_ev}
	\resizebox{\columnwidth}{!}{
		\begin{tabular}{clcccc}
			\toprule
			& & \multicolumn{2}{c}{Separately} & \multicolumn{2}{c}{Simultaneously} \\
			\cmidrule{3-6}
			& & Mean & Std & Mean & Std \\ \midrule
			\multirow{4}{*}{\makecell{Exp. 1\\\texttt{cartpole}}} & \texttt{cartpole}$(A)$ & 299.95 & 0.10 & 261.944 & 48.094 \\  
			& \texttt{cartpole}$(B)$ & 299.79 & 0.43 & 280.098 & 38.178 \\  
			& \texttt{cartpole}$(C)$ & 299.97 & 0.05 & 274.072 & 36.804 \\  
			& \texttt{cartpole}$(D)$ & 300.00 & 0.00 & 287.082 & 15.272 \\ 
			\midrule
			\multirow{4}{*}{\makecell{Exp. 2\\\texttt{acrobot}}} & \texttt{acrobot}$(A)$ & -77.75 & 1.13 & -82.732 & 2.651 \\ 
			& \texttt{acrobot}$(B)$ & -87.72 & 1.63 & -95.412 & 5.634 \\ 
			& \texttt{acrobot}$(C)$ & -96.91 & 3.51 & -101.836 & 4.588 \\  
			& \texttt{acrobot}$(D)$ & -103.07 & 2.32 & -112.96 & 4.861 \\ 
			\midrule
			\multirow{4}{*}{\makecell{Exp. 3\\\texttt{pendulum}}} & \texttt{pendulum}$(A)$ & -270.39 & 99.44 & -800.915 & 171.822 \\
			& \texttt{pendulum}$(B)$ & -174.63 & 19.54 & -562.826 & 138.927 \\ 
			& \texttt{pendulum}$(C)$ & -176.83 & 19.65 & -722.138 & 220.063 \\
			& \texttt{pendulum}$(D)$ & -161.58 & 9.81 & -516.174 & 37.71 \\ \bottomrule
		\end{tabular}%
	}
\end{table}

Table \ref{tab:model_ev} lists the results obtained for addressing RQ2, comprising three experiments (one per environment). By analyzing them, it can be claimed that MFEA allows evolving multiple DQL models efficiently by using the designed unified search space. When optimizing for the \texttt{cartpole} environment (Exp. 1), it can be seen that the DQL models evolved can perform competitively with their individually evolved counterparts, even if being granted the same computational budget. Likewise, the \texttt{acrobot} environment (Exp. 2) does not undergo a strong degradation either, reaching rewards close to those of the individually evolved models. However, in \texttt{pendulum} (Exp. 3) notably worse results were obtained. This fact unveils that the complexity of the environment to be solved directly impacts on the quality of the multifactorial evolution. The spotlight must be placed on the reasons for this degradation, reason for which an study on the transferability of knowledge among tasks (RQ3) is addressed in the next section.

\subsection*{RQ3: Can we quantify the intra- and inter-environment knowledge transfer among RL tasks promoted by MFEA?} \label{exp:genetic_transference}

We end our discussion on the performance of the proposed framework by delving into the genetic knowledge transfer of MFEA. In general, Transfer Learning is applied among Deep Neural Networks by copying weights and biases from a pre-trained network in a origin task $T_k$ to a network aimed to solve a target task $T_{k'}$. In MFEA, however, the knowledge transfer is done throughout the evolution process, sharing the knowledge acquired by the population and embedded in the genotype of the search space $\bm{\mathcal{X}}^U$. To undertake this knowledge transferability study, we set MFEA to evolve simultaneously $K=9$ tasks corresponding to $3$ different configurations (namely, $A$, $B$ and $C$ as per Table \ref{tab:new_envs}) of each of the $N=3$ environments considered in our experiments.  Once this experiment is designed and run, we record similar final reward statistics to those reported for RQ1 and RQ2. 
\begin{table}[h!]
	\vspace{-2mm}
	\centering
	\caption{Average and standard deviation of rewards obtained by MFEA when evolving jointly DQL models aimed at diverse RL tasks (intra- and inter-environment transfer)}
	\label{tab:allversions}
	\resizebox{\columnwidth}{!}{
		\begin{tabular}{ccccccc}
			\toprule
			& \multicolumn{2}{c}{$\alpha=A$} & \multicolumn{2}{c}{$\alpha=B$} & \multicolumn{2}{c}{$\alpha=C$} \\ \cmidrule{2-7}
			& Mean & Std & Mean & Std & Mean & Std \\ \midrule
			\texttt{cartpole}$(\alpha)$ & 229.21 & 57.6 & 273.87 & 32.25 & 218.82 & 57.21 \\ 
			\texttt{acrobot}$(\alpha)$ & -84.52 & 3.11 & -95.86 & 2.47 & -116.221 & 5.79 \\ 
			\texttt{pendulum}$(\alpha)$ & -1086.02 & 69.52 & -580.34 & 72.80 & -721.88 & 144.43 \\ \bottomrule
		\end{tabular}
	}
\end{table}

Results are listed in Table \ref{tab:allversions}, which must be analyzed by comparing them to those in Table \ref{tab:model_ev}. Relatively lower average reward values are reached for all the tasks when evolved jointly, with gaps being narrower for $\texttt{acrobot}$ tasks. Despite this worse performance, it is important to bear in mind that in this case, more -- and more diverse -- tasks are evolved simultaneously with the same evaluation budget and population size. This demonstrates the potential of MFEA to evolve DQL models for complicated RL tasks at the same time. On the other hand, it provides evidence that the MFEA operators devised to promote genetic transfer among tasks do not outweigh the limited computational budget allocated for all RL problems. It is appropriate to examine the reasons for this lack of effectiveness, and discriminate whether the joint evolution of tasks defined in these three environments undergoes negative information transfer among such tasks that cannot be evaded by MFEA. 
\begin{figure}[tbh!]
	\centering
	\vspace{-2mm}
	\includegraphics[width=0.9\columnwidth]{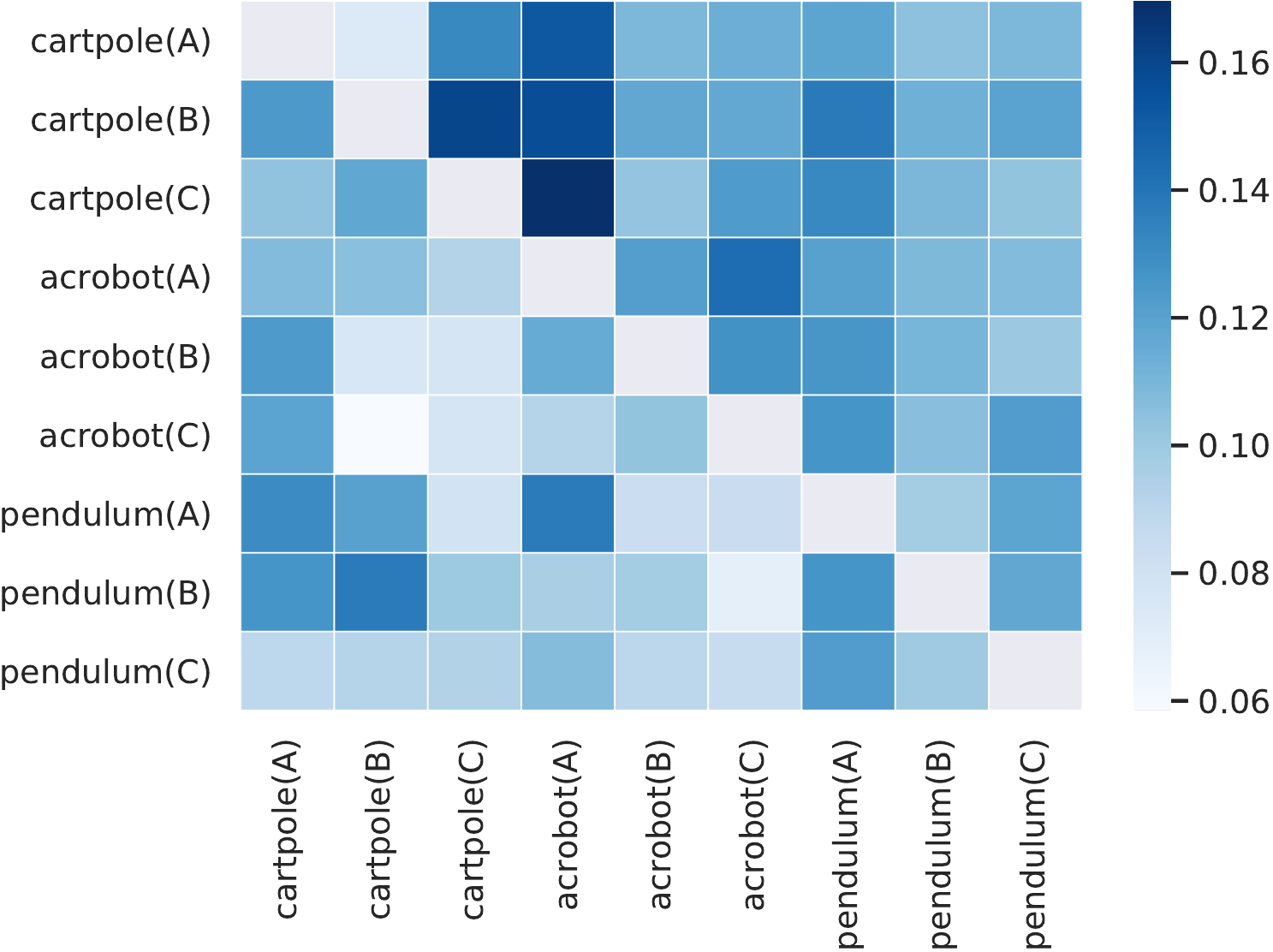}
	\vspace{-4mm}
	\caption{Effective crossover percentage between tasks, denoting the percentage of times that a crossover with the task denoted in the row has been effective for the task denoted in the column.}
	\label{fig:effective_cr}
\end{figure}

To this end, we check the times an \emph{effective crossover} has occurred among tasks, referring as such to the crossover event in which a candidate with skill factor equal to task $k$, when recombined with another candidate with skill factor ${k'}$, improves the performance of the latter in task $T_{k'}$. The heat map depicted in Figure \ref{fig:effective_cr} indicates the ratio of effective crossovers for every pair of tasks as per row and column labels. For instance, a highly intense color in the cell with coordinates (row,column)$=(T_k,T_{k'})$ indicates that the crossover operation between an individual with skill factor $k$, when mating with an individual with skill factor equal to ${k'}$, has positively contributed to solving $T_{k'}$. Interesting relationships emerge from this plot, such as a remarkably positive knowledge transfer between $\texttt{cartpole}(C)$ and $\texttt{acrobot}(A)$, followed by lower yet equally notable contributions from $\texttt{cartpole}(B)$ and $\texttt{cartpole}(A)$. On the contrary, tasks whose average reward as per Table \ref{tab:allversions} was found to be poor seem not to contribute significantly to the convergence of tasks defined on other environments (i.e. low inter-environment transfer), as can be read for e.g. $\texttt{pendulum}(C)$. However, for this latter example a higher intra-environment transfer is noted, with $\texttt{pendulum}(B)$ having received a positive push recurrently when mating with $\texttt{pendulum}(C)$.

\section{Conclusions and Future Research} \label{sec:conclusions}

This work has presented the application of Multfactorial Optimization to simultaneously optimize multiple DQL models devised for a set of interrelated RL tasks. The goal is to evolve jointly the weights and biases of the Deep Neural Networks inside the aforementioned DQL models by formulating it as a single optimization problem defined on an unified search space, whose design is inspired by the usual way of performing Transfer Learning among Deep Neural Networks. A framework relying on the so-called Multifactorial Evolutionary Algorithm (MFEA) has been designed to efficiently explore this unified search space, incorporating further operators to probabilistically exploit possible synergies among the tasks under consideration during the search process.

We have empirically gauged the performance of MFEA to solve multiple RL scenarios at the same time. MFEA has been shown to perform well for this scenario, evolving up to nine tasks by using a unique evolution process and just one population, which is shared between all the environments/tasks. We have also evaluated the effectiveness of the knowledge transfer when evolving DQL models, with interesting insights between the complexity of the RL task and its contribution in terms of knowledge transfer to the rest of the tasks. 

Future work will be devoted to the design of innovative knowledge sharing strategies, as well as new encoding schemes that extend the capabilities of the devised framework, i.e. evolution of larger networks and more complex environments. Efforts will be also invested towards optimizing DQL tasks by MFEA in an on-line fashion, i.e. while partially optimized DQL models interact with the environment.

\section*{Acknowledgments}

Aritz D. Martinez, Eneko Osaba and Javier Del Ser would like to thank the Basque Government for its funding support through the EMAITEK and ELKARTEK programs. Javier Del Ser receives funding support from the Consolidated Research Group MATHMODE (IT1294-19) granted by the Department of Education of the Basque Government.

\bibliography{bibliography}
\bibliographystyle{ieeetr}

\end{document}